\pdfoutput=1

\documentclass[sigconf]{acmart}
\usepackage{soul}
\usepackage{adjustbox}
\usepackage{pgfplots}

\AtBeginDocument{%
  \providecommand\BibTeX{{%
    \normalfont B\kern-0.5em{\scshape i\kern-0.25em b}\kern-0.8em\TeX}}}

\copyrightyear{2021}
\acmYear{2021}
\setcopyright{acmcopyright}
\acmConference[CIKM '21] {Proceedings of the 30th ACM International Conference on Information and Knowledge Management}{November 1--5, 2021}{Virtual Event, Australia.}
\acmBooktitle{Proceedings of the 30th ACM Int'l Conf. on Information and Knowledge Management (CIKM '21), November 1--5, 2021, Virtual Event, Australia}
\acmPrice{15.00}
\acmISBN{978-1-4503-8446-9/21/11}
\acmDOI{10.1145/XXXXXX.XXXXXX}
\usepackage{microtype}
\begin{document}
\fancyhead{}

\title{Aspect Sentiment Triplet Extraction Using\\ Reinforcement Learning}

\settopmatter{authorsperrow=4}

\author{Samson Yu Bai Jian}
\affiliation{%
  \institution{SUTD, Singapore}
  \streetaddress{8 Somapah Road}
  %\city{Singapore}
  %\state{Singapore}
  \country{}
  %\postcode{487372}
}
\email{samson_yu@sutd.edu.sg}

\author{Tapas Nayak}
\affiliation{%
  \institution{IIT KGP, India}
  \streetaddress{}
  %\city{Kharagpur}
  %\state{West Bengal}
  \country{}
 % \postcode{721302}
}
\email{tnk02.05@gmail.com}

\author{Navonil Majumder}
\affiliation{%
  \institution{SUTD, Singapore}
  %\streetaddress{8 Somapah Road}
 % \city{Singapore}
 % \state{Singapore}
 \country{}
  %\postcode{487372}
}
\email{navonil_majumder@sutd.edu.sg}

\author{Soujanya Poria}
\affiliation{%
 \institution{SUTD, Singapore}
% \streetaddress{8 Somapah Road}
% \city{Singapore}
% \state{Singapore}
 \country{}
% \postcode{487372}
}
\email{sporia@sutd.edu.sg}
% \author{Samson Yu Bai Jian$^+$, Tapas Nayak$^\dagger$, Navonil Majumder$^+$, Soujanya Poria$^+$}
% \affiliation{%
%   $^+$ ISTD, Singapore University of Technology and Design \country{Singapore} \\
%  % {\tt \{samson\_yu, navonil\_majumder, sporia\}@sutd.edu.sg } \\
%     $^\dagger$ Department of Computer Science \& Engineering,
%   Indian Institute of Technology Kharagpur \country{India} \\
%  % {\tt tnk02.05@gmail.com}}
\renewcommand{\shortauthors}{Yu et al.}
\newcommand{\mymodel}{ASTE-RL}

\begin{abstract}
Aspect Sentiment Triplet Extraction (ASTE) is the task of extracting triplets of aspect terms, their associated sentiments, and the opinion terms that provide evidence for the expressed sentiments. Previous approaches to ASTE usually simultaneously extract all three components or first identify the aspect and opinion terms, then pair them up to predict their sentiment polarities. In this work, we present a novel paradigm, \textbf{\mymodel{}}, by regarding the aspect and opinion terms as arguments of the expressed sentiment in a hierarchical reinforcement learning (RL) framework. We first focus on sentiments expressed in a sentence, then identify the target aspect and opinion terms for that sentiment. This takes into account the mutual interactions among the triplet's components while improving exploration and sample efficiency. Furthermore, this hierarchical RL setup enables us to deal with multiple and overlapping triplets. In our experiments, we evaluate our model on existing datasets from laptop and restaurant domains and show that it achieves state-of-the-art performance. The implementation of this work is publicly available at \url{https://github.com/declare-lab/ASTE-RL}.
\end{abstract}

%   <concept_id>10010147.10010178.10010179</concept_id>
%   <concept_desc>Computer methodologies~Natural language processing</concept_desc>
%   <concept_id>10010147.10010257</concept_id>
%   <concept_desc>Computer methodologies~Machine learning</concept_desc>

% \ccsdesc[500]{Computer methodologies~Natural language processing}
% \ccsdesc[300]{Computer methodologies~Machine learning}

% \keywords{aspect sentiment triplet extraction, sentiment analysis, neural networks, reinforcement learning, sequence labeling}

\maketitle

\section{Introduction}

% ACM's consolidated article template, introduced in 2017, provides a
% consistent \LaTeX\ style for use across ACM publications, and
% incorporates accessibility and metadata-extraction functionality
% necessary for future Digital Library endeavors. Numerous ACM and
% SIG-specific \LaTeX\ templates have been examined, and their unique
% features incorporated into this single new template.

% If you are new to publishing with ACM, this document is a valuable
% guide to the process of preparing your work for publication. If you
% have published with ACM before, this document provides insight and
% instruction into more recent changes to the article template.

% The ``\verb|acmart|'' document class can be used to prepare articles
% for any ACM publication --- conference or journal, and for any stage
% of publication, from review to final ``camera-ready'' copy, to the
% author's own version, with {\itshape very} few changes to the source.

Aspect-based sentiment analysis (ABSA) or target-based sentiment analysis (TBSA) is an important research area in natural language processing (NLP)~\cite{poria2020beneath, duan2021survey}. It consists of various fine-grained sentiment analysis tasks \cite{nasukawa2003sentiment, liu2010sentiment, liu2012sentiment}, with the three most fundamental being aspect/target term extraction, opinion term extraction and aspect/target term sentiment classification. Aspect Sentiment Triplet Extraction (ASTE) is a relatively new subtask of ABSA introduced by \citet{Li2019Unified,peng2020knowing}. In ASTE, the task is to extract triplets containing aspect terms, their associated sentiment polarities, and the opinion terms that express those sentiments. A sentence may contain multiple such triplets where the aspect terms or opinion terms across triplets may overlap with each other. We include an example of such triplets in Table \ref{tab:example}.

% Consider an example: "\emph{The pizza is the best if you like thin crusted pizza}". In this sentence, the aspect term is "\emph{pizza}", the opinion term is "\emph{best}" and there is the positive sentiment towards the aspect term. More complex tasks in ABSA are mostly a combination of these fundamental tasks.

% An ASTE model should be able to extract multiple triplets and overlapping triplets.

\begin{table}[ht!]
\centering
\small
\caption{An Example of ASTE Triplets Present in a Sentence}
\label{tab:example}
\resizebox{0.8\linewidth}{!}{
\begin{tabular}{|l|l|}
\hline
Sentence & \begin{tabular}[c]{@{}l@{}} Appetizers are excellent ; you can make a great \\( but slightly expensive ) meal out of them .\end{tabular}                                            \\ \hline
Triplets & \begin{tabular}[c]{@{}l@{}}{[}Aspect ; Opinion ; Sentiment{]}\\ (1) Appetizer ; excellent ; positive\\ (2) meal ; great ; positive\\ (3) meal ; slightly expensive ; negative\end{tabular} \\ \hline
\end{tabular}
}
\vspace{-1em}
\end{table}
Existing methods, such as, CMLA+ \cite{wang2017coupled}, RINANTE+ \cite{dai-song-2019-neural}, Li-unified-R \cite{Li2019Unified}, WhatHowWhy \cite{peng2020knowing}, OTE-MTL \cite{zhang2020multi}, GTS \cite{wu2020grid}, JET \cite{xu2020position}, TOP \cite{huang2021first} and BMRC \cite{chen2021bidirectional} are mainly divided into simultaneous and sequential methods. Early works \cite{wang2017coupled, dai-song-2019-neural, Li2019Unified, peng2020knowing, zhang2020multi, wu2020grid} usually employ a two-staged approach where they simultaneously extract aspect terms with sentiments and opinion terms. These triplets are subsequently decoded through triplet classification or pairwise matching. Recent works \cite{huang2021first, chen2021bidirectional} have shifted towards a more multi-stage, restrictive and sequential process during the extraction stage that can potentially capture more mutual dependencies and correlations among the triplet's components while forgoing the triplet decoding stage.

In this work, we tackle the ASTE task using a novel paradigm \textbf{\mymodel{}} where we consider the aspect and opinion terms as arguments of the sentiments expressed in a sentence. Previous approaches usually simultaneously extract all three components or first identify the aspect and opinion terms, then pair them up to predict their sentiment polarities. Unlike previous approaches, we propose a hierarchical reinforcement learning (RL) framework \cite{takanobu2019hierarchical, duan2020actionet} where we first consider the sentiment polarities, then identify their associated opinion and aspect terms using separate RL processes. This process is repeated to extract all triplets present in a sentence. With this hierarchical RL setup, the model handle multiple triplets and overlapping triplets, and model interactions between the three components effectively. Inspired by the recent success of the multi-turn machine reading comprehension (MRC) framework \cite{li-etal-2019-entity, chen2021bidirectional}, we incorporate ideas to further improve mutual interactions.

\section{Proposed Framework}
\subsection{Overview}

We divide our framework \textbf{\mymodel{}} into three components: 1) aspect-oriented sentiment classification, 2) opinion term extraction and 3) aspect term extraction. For the sentiment classification component, the sentiment is expressed towards the aspect term, and has four possible labels: $L^{C}=\{\mathrm{none, positive, negative, neutral}\}$. Our opinion and aspect extraction components are sequence labeling models with a BIO tagging scheme \cite{ramshaw1999text}. With this BIO scheme, we have three different labels to tag an input sequence for the opinion/aspect terms: $L^{O,A} = \{\mathrm{B,I,O}\}$. For a given sentence $X=\{x_1, x_2,..., x_J\}$ with $J$ tokens, \mymodel{} aims to output a set of labels $\{(Y^{O},Y^{A},C)^{|K|}\}$ where $|K|$ is the number of labels, $Y^{O}=\{y_1^O, y_2^O,..., y_J^O\}$ represents the tagging labels for the opinion term in a predicted triplet, $Y^{A}=\{y_1^A, y_2^A,..., y_J^A\}$ represents the tagging labels for the aspect term, and $C=\{c\}$ represents the sentiment polarity.

The three components are structured in a two-level hierarchy \citep{takanobu2019hierarchical}. In the higher level, we have the sentiment indicator. During the sequential scan of a sentence, an agent will decide at each position $j$ in a sentence at the token $x_j$ if it has gathered sufficient information to mark the position as indicative of a sentiment that is expressed towards an aspect term. If not, the agent will mark it as $\mathrm{none}$. Otherwise, it will mark it as either $\mathrm{positive, negative}$ or $\mathrm{neutral}$. In the latter case, the agent launches two subtasks in the lower level for the opinion and aspect extractions to identify the terms as arguments of the sentiment and engages in sequence labeling. Upon completion, the agent will return to the high-level sentiment indication process and continue the sequential scan of the sentence. This process is well-suited to be formulated as a semi-Markov decision process \citep{sutton1999between}: 1) a high-level RL process that detects a sentiment indicator in a sentence; 2) two low-level RL processes that identify the opinion and aspect terms separately for the corresponding sentiment.

\subsection{Aspect-Oriented Sentiment Classification with High-Level RL}
The high-level RL policy $\pi_u$ aims to detect the aspect-oriented sentiments in a sentence. This can be seen as a RL policy over options, where options are high-level actions \citep{sutton1999between}.

\textbf{Option:}
The option $\mathrm{\textbf{o}}_t$ is selected from $L^{C} = \{\mathrm{none, positive,}$ $\mathrm{negative, neutral}\}$ where $\mathrm{none}$ indicates no sentiment indicated towards any aspect term.

\textbf{State:}
The state $\mathrm{\textbf{s}}_t^u$ at each time step $t$ is represented by: 1) the current hidden state $\mathrm{\textbf{h}}_t$, 2) the current part-of-speech (POS) tag $\mathrm{\textbf{p}}_t$, 3) the sentiment polarity vector $\mathrm{\textbf{v}}_t^c$, and 4) the high-level state for the previous time step $\mathrm{\textbf{s}}_{t-1}^u$ . To obtain $\mathrm{\textbf{p}}_t$ for each token in a sentence, we pass the sentence into the spaCy (\url{https://spacy.io/}) POS tagger. The sentiment polarity vector $\mathrm{\textbf{v}}_t^c$ is the embedding of the latest option $\mathrm{\textbf{o}}_{t'}$ where $\mathrm{\textbf{o}}_{t'} \neq \mathrm{none}$. Both the POS tag and sentiment embeddings are learned parameters in the model. Hence, the state $\mathrm{\textbf{s}}_t^u$ is formally represented by: 
\begin{equation}
\mathrm{\textbf{s}}_t^u = f^{u}_{s}(\mathrm{\textbf{W}}_s^u[\mathrm{\textbf{h}}_t;\mathrm{\textbf{p}}_t;\mathrm{\textbf{v}}_t^c; \mathrm{\textbf{s}}_{t-1}^{u}]),
\end{equation}
where $f^{u}_{s}(\cdot)$ is a non-linear function implemented by a MLP. The hidden state $\mathrm{\textbf{h}}_t$ is obtained from a pre-trained BERT model \cite{devlin2018bert} with Whole Word Masking and fine-tuned on the SQuAD v1.1 training set \cite{rajpurkar2016squad}. Specifically, we first combine the query "\emph{Which tokens indicate sentiments relating pairs of aspect spans and opinion spans?}" and the review sentence $X$ into the BERT tokenizer to get a final input $I=\{[CLS],q^{u}_{1},...,q^{u}_{|q|},[SEP],x_1,x_2,...,x_J\}$. We then pass this input into the BERT model, and $\mathrm{\textbf{h}}_t$ represents the output vector from the BERT model that corresponds to the token $x_t$. The initial state $\mathrm{\textbf{s}}_{0}^u$ is initialized as: $\mathrm{\textbf{s}}_{0}^u = \textbf{0}$.
% \begin{equation}
% \mathrm{\textbf{s}}_{0}^u = \textbf{0}.
% \end{equation}

% BERT-Large-Uncased

\textbf{Policy:}
The stochastic policy for sentiment detection $\pi_u$ specifies a probability distribution over the options:
\begin{equation}
\mathrm{\textbf{o}}_t \sim \pi_u (\mathrm{\textbf{o}}_t|\mathrm{\textbf{s}}_t^u) = softmax(\mathrm{\textbf{W}}_{\pi}^{u}\mathrm{\textbf{s}}_t^u).
\end{equation}

\textbf{Reward:}
At every time step, when $\mathrm{\textbf{o}}_t$ is executed, the \textbf{intermediate reward} $\mathrm{\textbf{r}}_t^u$ provided by the environment follows this:
{%
\small
\begin{equation}
\mathrm{\textbf{r}}_{t}^u =
\begin{cases}
1,\quad\ \ \ \mathrm{if\ \textbf{o}_t\ in}\ X \\
0,\quad\ \ \ \mathrm{if\ \textbf{o}_t} = \mathrm{none} \\
-1,\quad\mathrm{if\ \textbf{o}_t\ not\ in}\ X.
\end{cases}
\end{equation}
}%

If a sentiment that is expressed towards an aspect term is detected at a time step (i.e. $\mathrm{\textbf{o}}_t \neq \mathrm{none}$), the agent will launch two subtasks as low-level RL processes. When the subtasks are completed, the agent will return to the high-level RL process. Otherwise, the agent continues its sequential scan of $X$ until the last option $\mathrm{\textbf{o}}_J$ about the last word $x_J$ of $X$ is sampled. When all options are sampled (i.e. at the end of the combined hierarchical RL process), there is a \textbf{final reward} $\mathrm{\textbf{r}}^u_{final}$ for the high-level process: $\mathrm{\textbf{r}}^u_{final} = F_1(X)$
% \begin{equation}
% \mathrm{\textbf{r}}^u_{final} = F_1(X)
% \end{equation}
where $F_1$ is the harmonic mean of the precision and recall in terms of the sentiment(s) in a sentence $X$.

\subsection{Opinion and Aspect Extractions with Low-Level RL}
Every time the high-level policy detects an aspect-oriented sentiment, two low-level policies $\pi_l^O,\pi_l^A$ will extract the corresponding opinion and aspect terms respectively and separately for the sentiment. In this subsection, we will generalize the RL elements such that they apply for both low-level RL processes, unless otherwise stated.

\textbf{Action:}
The action at every time step $t$ is to assign a tag to the current word. The action $\mathrm{\textbf{a}}_t$ is selected from $L^{O,A} = \{\mathrm{B,I,O}\}$, following a BIO tagging scheme. The $\mathrm{B/I}$ symbols represent the beginning and inside of an opinion/aspect term respectively, while the $\mathrm{O}$ symbol represents the unmarked label.

\textbf{State:}
Similar to the high-level policy, the state $\mathrm{\textbf{s}}_t^l$ at each time step $t$ is represented by: 1) the current hidden state $\mathrm{\textbf{h}}_t$, 2) the current POS tag $\mathrm{\textbf{p}}_t$, 3) the opinion/aspect tag vector $\mathrm{\textbf{v}}_t^{O,A}$, 4) the low-level state for the previous time step $\mathrm{\textbf{s}}_{t-1}^l$. To enhance the interactions between the sentiment and its associated opinion/aspect terms, we add a context vector $\mathrm{\textbf{v}}_{t'}^{ctx}$ to the state $\mathrm{\textbf{s}}_t^l$ at each time step $t$, using the sentiment state representation assigned to the latest option $\mathrm{\textbf{o}}_{t'}$:
\begin{equation}
\mathrm{\textbf{v}}_{t'}^{ctx} = g^{ctx}(\mathrm{\textbf{W}}^{ctx}_{v_{t'}}\mathrm{\textbf{s}}_{t'}^{u}).
\end{equation}
We also add the output vector from the BERT model for the [CLS] token, while computing the output vectors for the hidden states. Hence, the state $\mathrm{\textbf{s}}_t^l$ is formally represented by:
\begin{equation}
\mathrm{\textbf{s}}_t^l = f^{l}_{s}(\mathrm{\textbf{W}}_s^l[\mathrm{\textbf{h}}_t;\mathrm{\textbf{p}}_t;\mathrm{\textbf{v}}_t^{O,A}; \mathrm{\textbf{s}}_{t-1}^{l}; \mathrm{\textbf{v}}_{t'}^{ctx};\mathrm{\textbf{h}}_t^{CLS}]).
\end{equation}
Note that the representations used to compute the first low-level states for the opinion and aspect extractions are different. The representation used to compute the first low-level state for opinion term extraction $\mathrm{\textbf{s}}_1^{l,O}$ is initialized using $\mathrm{\textbf{s}}^u_{t'}$ as: $\mathrm{\textbf{s}}_{0}^{l,O} = g^l(\mathrm{\textbf{W}}^{l}_{s_{0}}\mathrm{\textbf{s}}_{t'}^u
)$.
% \begin{equation}
% \mathrm{\textbf{s}}_{0}^{l,O} = g^l(\mathrm{\textbf{W}}^{l}_{s_{0}}\mathrm{\textbf{s}}_{t'}^u
% ).
% \end{equation}
The representation used to compute the first low-level state for aspect term extraction $\mathrm{\textbf{s}}_1^{l,A}$ is initialized using $\mathrm{\textbf{s}}_J^{l,O}$ as: $\mathrm{\textbf{s}}_{0}^{l,A} = \mathrm{\textbf{s}}_J^{l,O}$.
% \begin{equation}
% \mathrm{\textbf{s}}_{0}^{l,A} = \mathrm{\textbf{s}}_J^{l,O}.
% \end{equation}
These initializations help us capture interactions between the triplet's components.
$f^{l}_{s}(\cdot)$ is a non-linear function implemented by a MLP, while $g^{ctx}(\cdot)$ and $g^{l}(\cdot)$ are linear functions that are implemented by a single linear layer. The hidden state $\mathrm{\textbf{h}}_t$ is obtained in the same way as in the high-level RL process. However, the queries are changed. The query is "\emph{What is the opinion span for the} $\mathrm{\textbf{o}}_{t'}$ \emph{sentiment indicated at} $x_{t'}$\emph{?}" for opinion term extraction and "\emph{What is the aspect span for the} $\mathrm{\textbf{o}}_{t'}$ \emph{sentiment indicated at} $x_{t'}$\emph{?}" for aspect term extraction.

\textbf{Policy:}
The stochastic policy for opinion/aspect extraction $\pi_l$ specifies a probability distribution over the actions given the low-level state $\mathrm{\textbf{s}}_t^l$ and the high-level option $\mathrm{\textbf{o}}_{t'}$ that launches the current subtask:
\begin{equation}
\mathrm{\textbf{a}}_t \sim \pi_l (\mathrm{\textbf{a}}_t|\mathrm{\textbf{s}}_t^l;\mathrm{\textbf{o}}_{t'}) = softmax(\mathrm{\textbf{W}}_{\pi}^{l}[\mathrm{\textbf{o}}_{t'}]\mathrm{\textbf{s}}_t^l).
\end{equation}

\textbf{Reward:}
At every time step, when $\mathrm{\textbf{a}}_t$ is executed, the \textbf{intermediate reward} $r^l_t$ is computed as the prediction error over the gold labels:
{%
\small
\begin{equation}
r^l_t = \begin{cases}
\lambda,\quad\ \ \ \ \ \ \mathrm{if\ tag} \ y_t\mathrm{\ is\ predicted\ correctly} \\
-0.5,\quad\mathrm{otherwise},
\end{cases}
\end{equation}
}%
where $\lambda \geq 0$ and $\lambda$ depends on the aspect/opinion tag type. This enables the model to learn a policy that emphasizes the prediction of B and I tags and avoids only predicting O tags in a trivial manner. When all actions are sampled, there is a \textbf{final reward} $r^l_{final}$ for the low-level processes, represented by:
{%
\small
\begin{equation}
r^l_{final} = \begin{cases}
1,\quad\ \ \ \mathrm{if\ all\ tags\ are\ predicted\ correctly} \\
-1,\quad\mathrm{otherwise}.
\end{cases}
\end{equation}
}%
There will also be negative rewards in the cases where the low-level processes produce impossible predictions, namely cases where there are no or more than one $\mathrm{B}$ tag present, and no or more than one opinion/aspect term identified for each predicted triplet. Note that the low-level rewards are non-zero only in the case where the option $\mathrm{\textbf{o}}_{t'}$ from the high-level process is correctly predicted.

\subsection{Hierarchical Policy Learning}
We learn the high-level policy by maximizing the expected total reward at each time step $t$ as the agent samples trajectories following the high-level policies $\pi_u$. Likewise, we learn the low-level policies by maximizing the expected total reward at each time step $t$ as the agent samples trajectories following the low-level policies $\pi_l^A, \pi_l^O$. We then optimize all policies using policy gradient methods \citep{sutton1999policy} with the REINFORCE algorithm \citep{williams1992simple, takanobu2019hierarchical}.

\subsection{Training Procedure}

We pre-train our \mymodel{} models for 40 epochs with a learning rate of 2e-5. During pre-training, we give our model the ground-truth options or actions at every time step to limit the exploration of the agent due to the high-dimensional state space in our setup. This prevents the agent from exploring too many unreasonable cases, e.g. an I tag preceding a B tag, and learning too slowly. We then fine-tune the best model (chosen based on the Dev $F_1$ score) with RL policy for 15 epochs with a learning rate of 5e-6. We sample 5 trajectories for each data point during RL fine-tuning. 

We initialize the BERT parameters from pre-trained weights \citep{devlin2018bert} and update them during training for this task. We set the dimension of sentiment polarity and opinion/aspect tag embeddings at 300. For POS embeddings, we set the dimension at 25. We randomly initialize these embeddings and update them during training. We set the state vector dimension for $\mathrm{\textbf{s}}^u_t$ and $\mathrm{\textbf{s}}^l_t$ at 300. We apply dropout \cite{srivastava2014dropout} after the non-linear activations in $f^{u}_{s}(\cdot)$ and $f^{l}_{s}(\cdot)$ during training and set the dropout rate at 0.5. We train our models in mini-batches of size 16 and optimize the model parameters using the Adam optimizer \cite{kingma2014adam}.

\section{Experiments and Analysis}

\subsection{Datasets \& Evaluation Metrics}

\begin{table}[ht]
\caption{ASTE-Data-V2 Dataset Statistics}
\label{tab:dataset_stat}
\centering
\resizebox{0.9\linewidth}{!}{
\begin{tabular}{l|ccc|ccc|ccc|ccc}
\toprule
           & \multicolumn{3}{c|}{\texttt{14Lap}} & \multicolumn{3}{c|}{\texttt{14Rest}} &
           \multicolumn{3}{c|}{\texttt{15Rest}} &
           \multicolumn{3}{c}{\texttt{16Rest}} \\ \hline
           & Train    & Dev    & Test   & Train    & Dev    & Test  
           & Train    & Dev    & Test   & Train    & Dev    & Test    \\
           \midrule
\#sentence & 906      & 219    & 328    & 1266     & 310    & 492  & 605	& 148	& 322 & 857	& 210	& 326  \\ 
\#positive & 817      & 169    & 364    & 1692     & 404    & 773   & 783	& 185	& 317	& 1015	& 252	& 407  \\ 
\#negative & 517      & 141    & 116    & 480      & 119    & 155  & 205	& 53	& 143	& 329	& 76	& 78   \\ 
\#neutral  & 126      & 36     & 63     & 166      & 54     & 66    & 25	& 11	& 25	& 50	& 11	& 29  \\
\bottomrule
\end{tabular}
}
\vspace{-1em}
\end{table}

We use the ASTE-Data-V2 dataset\footnote{https://github.com/xuuuluuu/SemEval-Triplet-data} curated by \citet{xu2020position} to show the effectiveness of \mymodel{} in two different domains of English reviews, namely the laptop and restaurant domains. \texttt{14Rest}, \texttt{15Rest}, \texttt{16Rest} are the datasets of the restaurant domain and \texttt{14Lap} is of the laptop domain. We include the statistics of the four datasets in ASTE-Data-V2 in Table \ref{tab:dataset_stat}, where \#sentence represents the number of sentences, and \#positive, \#negative, and \#neutral represent the numbers of triplets with positive, negative, and neutral sentiment polarities respectively.

We process the sentences with BERT's WordPiece tokenizer \cite{wu2016google} to make them work for \mymodel{}. Since the WordPiece tokenization may break down the tokens in the original dataset into subwords, we need to align the opinion/aspect term annotations and our BIO tagging scheme. We tag every token that corresponds to the opinion/aspect term tokens in the original annotations with $\mathrm{I}$, except for the first token, which we tag with $\mathrm{B}$. 

We follow the evaluation metrics of \citet{xu2020position} for our experiments. An extracted triplet is correct if the entire aspect term, opinion term, and sentiment polarity match with a ground-truth triplet. We report \textbf{precision}, \textbf{recall} and $\mathbf{F_1}$ score based on this.

% For example, if the original opinion/aspect annotation for a sentence $X_k=\{x_{1,k},x_{2,k},x_{3,k},x_{4,k}\}$ is $\{\mathrm{O,B,I,O}\}$ and the sentence is tokenized as $\{x_{1,k},x_{2,k},x_{2,k}^{sub},x_{3,k},x_{3,k}^{sub},x_{4,k}\}$ with the WordPiece tokenization, the new BIO annotation will be $\{\mathrm{O,B,I,I,I,O}\}$.

% \subsection{Evaluation Metrics}
% We follow the evaluation metrics of \citet{xu2020position} for our experiments. An extracted triplet is correct if the entire aspect term, opinion term, and sentiment polarity match with a ground-truth triplet. We report \textbf{precision}, \textbf{recall} and $\mathbf{F_1}$ score based on this.

\begin{table*}[ht!]
\small
\caption{Results of \mymodel{} and Previous Methods on the ASTE-Data-V2 Dataset}
\resizebox{0.8\linewidth}{!}
{
\label{tab:main_results}
\centering
\begin{tabular}{l|cccc|cccc|cccc|cccc}
\toprule
     & \multicolumn{4}{c|}{\texttt{14Lap}}                & \multicolumn{4}{c|}{\texttt{14Rest}}         
     & \multicolumn{4}{c|}{\texttt{15Rest}} 
     & \multicolumn{4}{c}{\texttt{16Rest}} \\ \hline
Model            & Dev $F_1$                & Prec                 & Rec                  & $F_1$                    & Dev $F_1$                & Prec                 & Rec                  & $F_1$             & Dev $F_1$                & Prec                 & Rec                  & $F_1$       & Dev $F_1$                & Prec                 & Rec                  & $F_1$              \\ \midrule
% CMLA+ \cite{wang2017coupled}            &          -             & 30.09                 & 36.92                 & 33.16                 &           -            & 39.18                 & 47.13                 & 42.79   & -   & 34.56	& 39.84	& 37.01  & - & 41.34	 & 42.10	& 41.72       \\ 
% RINANTE+ \cite{dai-song-2019-neural}          &            -           & 21.71                 & 18.66                 & 20.07                 &         -              & 31.42                 & 39.38                 & 34.95             & -     & 29.88	& 30.06	& 29.97 & - & 25.68	& 22.30	& 23.87\\ 
% Li-unified-R \cite{Li2019Unified}     &              -         & 40.56                 & 44.28                 & 42.34                 &              -         & 41.04                  & 67.35                 & 51.00            & -  & 44.72	& 51.39	& 47.82  & - & 37.33	& 54.51	& 44.31  \\ 
WhatHowWhy \cite{peng2020knowing} &            -           & 37.38                 & 50.38                 & 42.87                 &           -            & 43.24                 & 63.66                 & 51.46                 & - & 48.07	& 57.51	& 52.32 & - & 46.96	& 64.24	& 54.21\\
OTE-MTL \cite{zhang2020multi} & - & 54.26 & 41.07 & 46.75 & - & 63.07 & 58.25 & 60.56 & - & 60.88 & 42.68 & 50.18 & - & 65.65 & 54.28 & 59.42 \\
$\mathrm{GTS}_{\mathrm{BiLSTM}}$ \cite{wu2020grid} & - & 58.02 & 40.11 & 47.43 & - & 71.41 & 53.00 & 60.84 & - & 64.57 & 44.33 & 52.57 & - & 70.17 & 55.95 & 62.26 \\
$\mathrm{JET}_{\mathrm{BiLSTM}}^{t}$ \cite{xu2020position}             &  48.26  & 54.84                  & 34.44                 & 42.31                 & 53.14 & 66.76                 & 49.09                 & 56.58      & 55.06 & 59.77 & 42.27 & 49.52         & 58.45 & 63.59 & 50.97 & 56.59 \\ 
$\mathrm{JET}_{\mathrm{BiLSTM}}^{o}$ \cite{xu2020position}             &  45.83  & 55.98                  & 35.36                 & 43.34                 & 53.54 & 61.50                 & 55.13                 & 58.14   & 60.97   & 64.37	& 44.33	& 52.50       & 60.90  & 70.94	& 57.00	& 63.21 \\
\hline
$\mathrm{GTS}_{\mathrm{BERT}}$ \cite{wu2020grid} & - & 57.12 & 53.42 & 55.21 & - & 71.76 & 59.09 & 64.81 & - & 54.71 & 55.05 & 54.88 & - & 65.89 & 66.27 & 66.08 \\
$\mathrm{JET}_{\mathrm{BERT}}^{t}$ \cite{xu2020position} & 50.40 & 53.53 & 43.28 & 47.86 & 56.00 & 63.44 & 54.12 & 58.41 & 59.86 & 68.20 & 42.89 & 52.66 & 60.67 & 65.28 & 51.95 & 57.85 \\
$\mathrm{JET}_{\mathrm{BERT}}^{o}$ \cite{xu2020position} & 48.84 & 55.39 & 47.33 & 51.04 & 56.89 & 70.56 & 55.94 & 62.40 & 64.78 & 64.45 & 51.96 & 57.53 & 63.75 & 70.42 & 58.37 & 63.83 \\
TOP \cite{huang2021first} & - & 57.84 & 59.33 & 58.58 & - & 63.59 & 73.44 & 68.16 & - & 54.53 & 63.30 & 58.59 & - & 63.57 & 71.98 & 67.52 \\
BMRC \cite{chen2021bidirectional} & 56.08 & 65.91 & 52.15 & 58.18 & 62.83 & 72.17 & 65.43 & 68.64 & 72.47 & 62.48 & 55.55 & 58.79 & 70.91 & 69.87 & 65.68 & 67.35 \\
\hline
% \mymodel{} (pre-train only)        & 57.35  & 62.00 & 55.84 &  58.73 &  64.50 & 69.70 & 69.23 & 69.47 & 72.84 & 63.31 & 61.61 & 62.44 & 71.50 & 64.76 & 70.74 & 67.57  \\
\textbf{\mymodel{}}   & 58.14 & 64.80  & 54.99 &  \textbf{59.50}   &  64.40  & 70.60   & 68.65 &  \textbf{69.61}  & 74.01 & 65.45 & 60.29 &  \textbf{62.72}  & 72.11 & 67.21 & 69.69 &  \textbf{68.41} \\
 \hspace{6mm} - Pre-training only        & 57.35  & 62.00 & 55.84 &  58.73 &  64.50 & 69.70 & 69.23 & 69.47 & 72.84 & 63.31 & 61.61 & 62.44 & 71.50 & 64.76 & 70.74 & 67.57  \\
\bottomrule
\end{tabular}
}
\end{table*}

\subsection{Baselines}

% We compare the performance of \mymodel{} against the following baselines: (i) \textbf{CMLA+}: \citet{wang2017coupled} proposed CMLA, a multi-layer attention model to capture the relationship between words and co-extract aspect and opinion terms. \citet{peng2020knowing} adapts CMLA for ASTE by enabling it to find the sentiment polarity too. (ii) \textbf{RINANTE+}: \citet{dai-song-2019-neural} proposed RINANTE, a LSTM-CRF \cite{lample2016neural} model to extract aspect and opinion terms. \citet{peng2020knowing} adapts RINANTE for ASTE using the same method as CMLA+. (iii) \textbf{Li-unified-R}: \citet{Li2019Unified} proposed Li-unified, a multi-layer LSTM neural architecture for co-extraction of aspect terms with sentiments. \citet{peng2020knowing} adapts Li-unified for ASTE by enabling opinion term extraction, and applying the same classifier as CMLA+.

We compare the performance of \mymodel{} against the following baselines: (i) \textbf{WhatHowWhy}: \citet{peng2020knowing} proposed a multi-layer LSTM neural architecture for co-extraction of aspect terms with sentiments, and opinion terms, with a Graph Convolutional Network \cite{kipf2016semi} component to capture dependency information to enhance the co-extraction. (ii) \textbf{OTE-MTL}: \citet{zhang2020multi} proposed a multi-task learning framework to jointly extract aspect and opinion terms while parsing word-level sentiment dependencies, before conducting a triplet decoding process. We use results from \citet{huang2021first} for OTE-MTL's performance on ASTE-Data-V2. (iii) \textbf{GTS}: \citet{wu2020grid} proposed an end-to-end grid tagging framework and a grid inference strategy to exploit mutual indication between opinion factors. We use results from \citet{huang2021first} for GTS' performance on ASTE-Data-V2, and report them for two variants: bidirectional LSTM (BiLSTM) and BERT. (iv) \textbf{JET}: \citet{xu2020position} proposed a position-aware tagging scheme for triplet extraction. They encode information about sentiment polarities and distances between the start position of aspect term and the opinion term's start and end positions ($\mathrm{JET}^{t}$) or vice versa ($\mathrm{JET}^{o}$). We report the results for two variants: BiLSTM and BERT. (v) \textbf{TOP}: \citet{huang2021first} proposed a two-stage method to enhance correlations between aspect and opinion terms. Aspect and opinion terms are first extracted with sequence labeling, and artificial tags are added to each pair to establish correlation. A sentiment polarity is then identified for each pair using the resulting representations. (vi) \textbf{BMRC}: \citet{chen2021bidirectional} proposed a transformation of the ASTE task into a multi-turn MRC task and a bidirectional MRC framework to address it. They use non-restrictive, restrictive and sentiment classification queries in a three-turn process to extract triplets. We train and test BMRC on ASTE-Data-V2 over 5 runs with different random seeds.

% \subsection{Experimental Settings}

% The BERT model is trainable. The sentiment polarity, opinion/aspect tag and POS tag embeddings are randomly initialized and have trainable parameters. The sentiment polarity and opinion/aspect tag embeddings have a dimensionality of 300, while the POS tag embeddings have a dimensionality of 25. The state vector dimension for $\mathrm{\textbf{s}}^u_t$ and $\mathrm{\textbf{s}}^l_t$ is 300. A dropout \cite{srivastava2014dropout} rate of 0.5 is applied after the non-linear activations in $f^{u}_{s}(\cdot)$ and $f^{l}_{s}(\cdot)$ during training. All models are optimized with the Adam optimizer \cite{kingma2014adam} and a batch size of 16.

% We pre-train our models for 40 epochs with a learning rate of 2e-5. During pre-training, we give \mymodel{} ground-truth options or actions at every time step to limit the exploration of the agent due to the high-dimensional state space in our setup. This prevents the agent from exploring too many unreasonable cases, e.g. an I tag preceding a B tag, and learning too slowly. We then fine-tune the best model (chosen based on the Dev $F_1$ score) with RL policy for 15 epochs with a learning rate of 5e-6. We sample 5 trajectories for each data point during RL fine-tuning. 

% the best models (as determined by their Dev $F_1$ scores)
\begin{table}[ht!]
\caption{$F_1$ Scores for Multiple and Overlapping Triplets}
\small
\label{tab:multiple_triplets}
\centering
\resizebox{0.9\linewidth}{!}{
\begin{tabular}{l|cc|cc|cc|cc}
\toprule
           & \multicolumn{2}{c|}{\texttt{14Lap}} & \multicolumn{2}{c|}{\texttt{14Rest}} &
           \multicolumn{2}{c|}{\texttt{15Rest}} &
           \multicolumn{2}{c}{\texttt{16Rest}} \\ \hline
           & ASTE-RL & BMRC & ASTE-RL & BMRC & ASTE-RL & BMRC & ASTE-RL & BMRC
           \\
           \midrule
Single & 62.46  & 63.04 & 67.44   & 66.51 & 61.33  & 59.59 & 66.31 & 67.83 \\
Multiple & 57.70 & 53.77 & 70.23 & 69.17 & 63.95 & 56.37 & 69.84 & 64.11 \\ \hline
No Overlap & 62.24  & 62.80 & 72.16   & 70.81 & 62.17  & 59.75 & 70.13 & 69.23 \\
Overlap & 55.94  & 50.18 & 67.23    & 63.96 & 63.83  & 55.48 & 65.20 &  58.59\\
\bottomrule
\end{tabular}
}
\vspace{-2.4em}
\end{table}
\subsection{Experimental Results}

The experimental results are shown in Table \ref{tab:main_results}. We observe that BERT-based models (their results are in the row above \mymodel{}'s results in Table \ref{tab:main_results}) generally perform better than the non-BERT models. Hence, we only experiment with BERT for our \mymodel{} model. We select our best model for each dataset based on its Dev $F_1$ score. For reproducibility, we report the testing results averaged over 5 runs with different random seeds. \mymodel{} outperforms existing baselines on all four datasets, and significantly outperforms existing baselines on the \texttt{15Rest} dataset. When compared to the second-best performance for each dataset, we observe an average improvement of 1.68\% $F_1$ score across all four datasets, and an improvement of 3.93\% on \texttt{15Rest}. We also observe that our model strikes a balance between the TOP and BMRC models in terms of precision and recall, and hypothesize that this balance can be flexibly shifted depending on $\beta$ to fit dataset requirements, if we generalize $\mathrm{\textbf{r}}^u_{final} = F_{\beta}(X)$, where $F_{\beta}(X)$ is the weighted harmonic mean of precision and recall.

\subsection{Effect of RL Fine-tuning}

In Table \ref{tab:main_results}, we report our results for \mymodel{} without the RL fine tuning step. In this setting, we pre-train our \mymodel{} for 40 epochs as usual and after that we run for another 15 epochs with a learning rate of 5e-6 (as used in RL fine-tuning step). As compared to the RL fine-tuning setting with multinomial sampling, this setting has lower $F_1$ scores with an average decrease of 0.51\% over 5 runs with different random seeds. In this setting, our model achieves slightly higher recall, but precision is significantly lower across all four datasets. This might be because multinomial sampling encourages more exploration after the initial pre-training of 40 epochs.

\vspace{-1em}
\subsection{Analysis on Multiple \& Overlapping Triplet Extraction}

We show the results of \mymodel{} and BMRC in complex situations where there are multiple and overlapping triplets in a sentence in Table \ref{tab:multiple_triplets}. For the multiple triplet scenario, we observe that there is a performance increase for \texttt{14Rest}, \texttt{15Rest} and \texttt{16Rest} and a decrease for \texttt{14Lap} as compared to the case where only one triplet is present in a sentence. For the overlapping triplet scenario, we observe a performance increase for for \texttt{15Rest} and a decrease for \texttt{14Lap}, \texttt{14Rest} and \texttt{16Rest}.

In general, we observe that \mymodel{} can handle multiple and overlapping triplets in a sentence consistently well due to its hierarchical RL setup, as compared to BMRC. There is a total $F_1$ decrease of 4.76\% for multiple triplet extraction for \mymodel{} across all four datasets as compared to 16.21\% for BMRC, and a total $F_1$ decrease for overlapping triplet extraction of 16.16\% for \mymodel{} as compared to 34.38\% for BMRC. 

\vspace{-0.7em}
\section{Conclusion}
In this work, we propose a novel \mymodel{} model based on hierarchical reinforcement learning (RL) paradigm for aspect sentiment triplet extraction (ASTE). In this paradigm, we treat the aspect and opinion terms as arguments of the sentiment polarities. We decompose the ASTE task into a hierarchy of three subtasks: high-level sentiment polarity extraction, and low-level opinion and aspect term extractions. This approach is good at modeling the interactions between the three tasks and handling multiple and overlapping triplets. We incorporate the multi-turn MRC elements in our model to further improve these interactions. Our proposed model achieves state-of-the-art performance on four challenging datasets for the ASTE task.

\section*{Acknowledgments}
%This project is supported by the TL-SEED grant at SUTD no. RTDSS1\newline910021 titled ``AI for Hate Speech Detection on Social Media'', the SRG grant at SUTD no. T1SRIS19149 titled ``An Affective Multimodal Dialogue System'', and grant no. RTDST190702 titled Complex QA.

This project is supported by the DSO grant no. RTDST190702  awarded to SUTD titled Complex Question Answering.

\bibliographystyle{ACM-Reference-Format}
\bibliography{sample-base}

%
% If your work has an appendix, this is the place to put it.
% \appendix

% \section{Rights Information}

% Authors of any work published by ACM will need to complete a rights
% form. Depending on the kind of work, and the rights management choice
% made by the author, this may be copyright transfer, permission,
% license, or an OA (open access) agreement.

% Regardless of the rights management choice, the author will receive a
% copy of the completed rights form once it has been submitted. This
% form contains \LaTeX\ commands that must be copied into the source
% document. When the document source is compiled, these commands and
% their parameters add formatted text to several areas of the final
% document:
% \begin{itemize}
% \item the ``ACM Reference Format'' text on the first page.
% \item the ``rights management'' text on the first page.
% \item the conference information in the page header(s).
% \end{itemize}

\end{document}